\documentclass[a4paper]{article}
\usepackage{aaai}
\usepackage{times}
\usepackage{helvet}
\usepackage{courier}

\nocopyright

\usepackage[utf8]{inputenc}
\usepackage[square]{natbib}
\usepackage[english]{babel}

\usepackage[font=it]{caption}

\usepackage{graphicx}
\usepackage{latexsym}

\usepackage{verbatim}
\usepackage{amssymb}
\usepackage{amsmath}

\usepackage{tikz,pgf}
\usetikzlibrary{arrows,backgrounds}
\usepackage{gnuplot-lua-tikz}
\usepackage{relsize}

\usepackage{url}
\usepackage[pdftex,colorlinks=true,linkcolor=black,citecolor=black,menucolor=black,urlcolor=black]{hyperref}

\newtheorem{theorem}{Theorem}
\newtheorem{proposition}{Proposition}
\newtheorem{example}{Example}

\newenvironment{proof}{\noindent{\bf Proof}}{\rule{2mm}{2mm}\\[-.5em] }

\newcommand{\Not}{not \,}

\newcommand{\Th}{T\!h}
\newcommand{\Atom}{Atom}
\newcommand{\Lit}{Lit}

\newcommand{\C}{{\mathbf{C}}}
\newcommand{\K}{{\mathbf{K}}}
\newcommand{\A}{{\mathbf{A}}}

\newcommand{\tr}{\mathit{tr}}

\newcommand{\Katom}{\Atom_\K}
\newcommand{\Aatom}{\Atom_\A}
\newcommand{\katom}[1]{k_{#1}}
\newcommand{\aatom}[1]{a_{#1}}

\newcommand{\trne}{\tr_{\mathit{ne}}}
\newcommand{\trlp}{\tr_{\mathit{lp}}}
\newcommand{\trc}{\tr_c}

\usepackage{color}

\newcommand{\set}[1]{\left\{#1\right\}}
\newcommand{\guard}{\ \middle\vert\ }

\def\limplies{\supset}

\newcommand{\theory}{\mathit{Th}}

\newcommand{\pre}{\phi}
\newcommand{\jus}{\psi}
\newcommand{\con}{\varphi}

\newcommand{\define}[1]{\emph{#1}}

\newcommand{\tdefault}[3]{{#1:#2/#3}}

\newcommand{\gkdlp}{{\textsf{gk2dlp}}}

\newcommand{\myparagraph}[1]{\vspace*{-1em}\paragraph{#1}}

\newenvironment{firstexample}[2]{
  \newcounter{#1:sec}
  \setcounter{#1:sec}{\value{section}}
  \newcounter{#1}
  \setcounter{#1}{\value{example}}
  \begin{example}[#2]
  \label{#1}
}{
  \end{example}
}

\newcounter{cexample}
\newcounter{csection}
\newenvironment{cexample}[1]{
  \setcounter{csection}{\value{section}}
  \setcounter{cexample}{\value{example}}
  \setcounter{section}{\value{#1:sec}}
  \setcounter{example}{\value{#1}}
  \begin{example}[Continued]
}{
  \end{example}
  \setcounter{section}{\value{csection}}
  \setcounter{example}{\value{cexample}}
}

\newcounter{trline}
\newcommand{\linenr}{{\stepcounter{trline}(\arabic{trline})}}

\newcommand{\sound}{\mathit{snd}}
\newcommand{\witness}{\mathit{wit}}
\newcommand{\formOne}{\Phi}
\newcommand{\formTwo}{\Psi}

\usepackage{ifthen}
\usepackage{environ}
\newif\iflong
\NewEnviron{shortproof}{\iflong\else\expandafter\begin{proof}\BODY\end{proof}\fi}
\NewEnviron{longproof}{\iflong\expandafter\begin{proof}\BODY\end{proof}\fi}
\longtrue

\newif\iffinal
\finaltrue

\newcommand{\presection}{}

\begin{document}

\title{Implementing Default and Autoepistemic Logics via the Logic of GK}
\author{%
  Jianmin Ji\\
  School of Computer Science and Technology\\
  University of Science and Technology of China\\
  Hefei, China
  \And
  Hannes Strass\\
  Computer Science Institute\\
  Leipzig University\\
  Leipzig, Germany
}

\maketitle

\begin{abstract}
The logic of knowledge and justified assumptions, also known as the logic of grounded knowledge (GK), was proposed by Lin and Shoham as a general logic for nonmonotonic reasoning. To date, it has been used to embed in it default logic (propositional case), autoepistemic logic, Turner's logic of universal causation, and general logic programming under stable model semantics. Besides showing the generality of GK as a logic for nonmonotonic reasoning, these embeddings shed light on the relationships among these other logics.
In this paper, for the first time, we show how the logic of GK can be embedded into disjunctive logic programming in a polynomial but non-modular translation with new variables. The result can then be used to compute the extension/expansion semantics of default logic, autoepistemic logic and Turner's logic of universal causation by disjunctive ASP solvers such as GNT, cmodels, DLV, and claspD(-2).
\end{abstract}



\section{Introduction}
\citet{gk} proposed a logic with two modal operators $\K$ and $\A$, standing for knowledge and assumption, respectively.
The idea is that one starts with a set of assumptions (those true under the modal operator $\A$), computes the minimal knowledge under this set of assumptions, and then
checks to see if the assumptions were justified in that they agree with
the resulting minimal knowledge.
For instance, consider the GK formula $\A p\supset \K p$.
If we assume $p$, then we can conclude that we know $p$, thus the assumption that $p$ holds is justified, and we get a GK model where both $\A p$ and $\K p$ are true.
(There is another GK model where we do not assume $p$ and hence do not know $p$.)
However, there is no GK model of $\neg \A p\supset \K p$:
if we do not assume $p$, we are forced to conclude $\K p$, but then knowledge and assumptions do not coincide;
if we do assume $p$, we cannot conclude that we know $p$ and thus assuming $p$ was not justified.

To date, there have been embeddings from default logic~\citep{reiter80} and
autoepistemic logic~\citep{Moore} to the logic of GK~\citep{gk}, from Turner's logic of universal causation~\citep{Turner} to the logic of GK~\citep{ji12}, as well as from general logic programs~\citep{Ferraris2005} to the logic of GK~\citep{LinZhou11}.
Among other things, these embeddings shed new light on nonmonotonic reasoning,
and have led to an interesting characterization of strong equivalence in
logic programming \citep{Lin:kr02,LinZhou11}, and helped relate logic
programming to circumscription \citep{gk} as the semantics of GK is just
a minimization (of knowledge) together with an identity check (of assumptions and knowledge) after the minimization.

In this paper, for the first time, we consider computing models of GK theories by disjunctive logic programs.
We shall propose a polynomial translation from a (pure) GK theory to a disjunctive logic program such that there is a one-to-one correspondence between GK models of the GK theory and answer sets of the resulting disjunctive logic program.
The result can then be used to compute the extension/expansion semantics of default logic, autoepistemic logic and Turner's logic of universal causation by disjunctive ASP solvers such as 
GNT~\citep{janhunen-gnt},
cmodels~\citep{cmodels},
DLV~\citep{dlv}, 
claspD~\citep{claspD} and
claspD-2~\citep{gebser13claspD2}.
In particular, the recent advances in disjunctive answer set solving~\citep{gebser13claspD2} open up promising research avenues towards applications of expressive nonmonotonic knowledge representation languages.

To substantiate this claim, we have implemented the translation and report on some preliminary experiments that we conducted on the special case of computing extensions for Reiter's default logic~\citep{reiter80}.
The implementation, called \gkdlp, is available for download from the second author's home page.\footnote{\url{http://informatik.uni-leipzig.de/~strass/gk2dlp/}}

Providing implementations for theoretical formalisms has a long tradition in nonmonotonic reasoning, for an overview see \citep{dix01nonmonotonic}.
In fact, nonmonotonic reasoning itself originated from a desire to more accurately model the way humans reason, and was since its conception driven by applications in commonsense reasoning~\citep{McCarthy80,McCarthy86}.
Today, thanks to extensive research efforts, we know how closely interrelated the different formalisms for nonmonotonic reasoning are, and can use this knowledge to improve the scope of implementations.

This paper is organized as follows. Section 2 reviews logic programs, the logic of GK and default and autoepistemic logics.
Section 3 presents our main result, the mapping from GK to disjunctive logic programming\iflong{}\else{; due to space constraints, we could however not include any of the proofs}\fi.
Section 4 presents our prototypical implementation, several experiments we conducted to analyze the translation, possible applications for it, and a comparison with previous and related work.
Section 5 concludes with ideas for future work.

\presection
\section{Preliminaries}

We assume a propositional language with two zero-place logical connectives $\top$ for tautology and $\bot$ for contradiction.
We denote by $\Atom$ the set of atoms,
the signature of our language, and
$\Lit$ the set of literals: $\Lit =\Atom\cup \{\neg p\mid p\in \Atom\}$.
A set $I$ of literals is called {\em complete} if for each atom $p$, exactly
one of $\{p,\neg p\}$ is in $I$.

In this paper, we identify an interpretation with a complete set of literals.
If $I$ is a complete set of literals, we use it as an interpretation
when we say that it is a model of a formula, and we use it as a set of literals
when we say that it entails a formula.
In particular, we denote by
$\Th(I)$ the logical closure of $I$ (considered to be a set of literals).


\presection
\subsection{Logic Programming}

A {\em nested expression} is built from literals using the 0-place connectives $\top$ and $\bot$, the unary connective ``$not$'' and the binary connectives ``,'' and ``;'' for conjunction and disjunction.
A {\em logic program} with nested expressions is a finite set of rules of the form
$F\gets G$, where $F$ and $G$ are nested expressions.
The \textit{answer set} of a logic program with nested expressions is defined as in~\citep{nested99}.
Given a nested expression $F$ and a set $S$ of literals, we define when $S$ satisfies $F$, written $S\models F$ below, recursively as follows ($l$ is a literal): 
\begin{itemize}
\item $S\models l$ if $l\in S$,
\item $S\models \top$ and $S\not\models \bot$,
\item $S\models \Not F$ if $S\not\models F$,
\item $S\models F, G$ if $S\models F$ and $S\models G$, and
\item $S\models F; G$ if $S\models F$ or $S\models G$.
\end{itemize}
$S$ satisfies a rule $F\gets G$ if $S\models F$ whenever $S\models G$.
$S$ satisfies a logic program $P$, written $S\models P$, if $S$ satisfies all rules in $P$.

The {\em reduct} $P^S$ of $P$ related to $S$ is the result of replacing every maximal subexpression of~$P$ that has the form $\Not F$ with $\bot$ if $S\models F$, and with $\top$ otherwise.
For a logic program $P$ without $not$, the {\em answer set} of $P$ is any minimal consistent subset $S$ of $\Lit$ that satisfies $P$.
We use $\Gamma_P(S)$ to denote the set of answer sets of $P^S$. Now a consistent set~$S$ of
literals is an {\em answer set} of~$P$ if{f}~$S \in \Gamma_P(S)$.
Every logic program with nested expressions can be equivalently translated to disjunctive logic programs with  disjunctive rules of the form
\begin{align*}
l_1;\cdots; l_k\gets &   l_{k+1}, \ldots, l_t, \Not  l_{t+1}, \ldots, \Not l_{m}, \\
&\Not\Not l_{m+1}, \ldots,\Not\Not l_n
\end{align*}
where $n\geq m \geq t\geq k \geq 0$ and $l_1, \ldots, l_n$ are propositional literals.

\presection
\subsection{Default Logic}
Default logic~\citep{reiter80} is for making and withdrawing assumptions in the light of incomplete knowledge.
This is done by \emph{defaults}, that allow to express rules of thumb such as ``birds usually fly'' and ``tools usually work.''
For a given logical language, a \define{default} is any expression of the form
\mbox{$\tdefault{\pre}{\jus_1, \ldots, \jus_n}{\con}$}
where $\pre, \jus_1, \ldots, \jus_n, \con$ are formulas of the underlying language.
A \define{default theory} is a pair $(W, D)$, where $W$ is a set of formulas and $D$ is a set of defaults.
The meaning of default theories is given through the notion of \emph{extensions}.
An extension of a default theory $(W, D)$ is ``interpreted as an acceptable set of beliefs that one may hold about the incompletely specified world $W$''~\citep{reiter80}.
For a default theory $(W, D)$ and any set $S$ of formulas let $\Gamma(S)$ be the smallest set satisfying
(1) $W \subseteq \Gamma(S)$,
(2) $\theory(\Gamma(S)) = \Gamma(S)$,
(3) If $\tdefault{\pre}{\jus_1, \ldots, \jus_n}{\con} \in D$, $\pre \in \Gamma(S)$ and $\neg \jus_1, \ldots, \neg \jus_n \notin S$, then $\con \in \Gamma(S)$.
A set $E$ of formulas is called an \define{extension for $(W, D)$} iff $\Gamma(E) = E$.

\presection
\subsection{Autoepistemic Logic}
\citet{Moore} strives to formalize an ideally rational agent reasoning about its own beliefs.
He uses a belief modality ${L}$ to explicitly refer to the agent's belief within the language.
Given a set $A$ of formulas (the initial beliefs), a set $T$ is an \define{expansion} of $A$ if it coincides with the deductive closure of the set
\mbox{$A \cup \set{ {L}\varphi \guard \varphi \in T } \cup \set{ \neg{L}\varphi \guard \varphi \notin T }$}.
In words, $T$ is an expansion if it equals what can be derived using the initial beliefs $A$ and positive and negative introspection with respect to $T$ itself.
It was later discovered that this definition of expansions allows unfounded, self-justifying beliefs.
Such beliefs are however not always desirable when representing the knowledge of agents.

\presection
\subsection{The Logic of GK}
\label{sec:gk}
The language of GK proposed by \citet{gk} is a modal propositional language with two modal operators, $\K$, for knowledge, and $\A$, for assumption.
GK {\em formulas} $\varphi$ are propositional formulas with $\K$ and $\A$, that is,
\begin{gather*}
  \varphi ::= \bot \mid p \mid \neg\varphi \mid \varphi\land\varphi \mid \varphi\lor\varphi \mid \K\varphi \mid \A\varphi
\end{gather*}
where $p$ is an atom.
A GK {\em theory} is a set of GK formulas.

GK is a nonmonotonic logic, and its semantics is defined using the
standard Kripke possible world interpretations. Informally speaking,
a GK model is a Kripke interpretation where what is true under $\K$ is
minimal and exactly the same as what is true under $\A$. The intuition
here is that given a GK formula, one first makes some assumptions (those true
under $\A$), then one minimizes the knowledge thus entailed, and finally
checks to make sure that the initial assumption is justified in the
sense that the minimal knowledge is the same as the initial assumption.

Formally,
a {\em Kripke interpretation}~$M$ is a tuple $\langle W, \pi, R_K, R_A, s\rangle$, where $W$ is a nonempty set of {\em possible worlds},
$\pi$ a function that maps a possible world to an interpretation, $R_K$ and $R_A$ binary relations over $W$ representing the accessibility relations for $\K$ and $\A$, respectively, and \mbox{$s\in W$}, called the {\em actual world} of $M$.
The {\em satisfaction relation} $\models$ between a Kripke interpretation \mbox{$M = \langle W, \pi, R_K, R_A, s\rangle$} and
a GK formula $\varphi$ is defined in a standard way:
\begin{itemize}
\item $M\not\models \bot$,
\item $M\models p$  if{f} $p \in \pi(s)$, where $p$ is an atom,
\item $M\models \neg \varphi$ if{f} $M\not\models \varphi$,
\item $M\models \varphi\land\psi$ if{f} $M\models \varphi$ and $M\models\psi$,
\item $M\models \varphi\lor\psi$ if{f} $M\models\varphi$ or $M\models\psi$,
\item $M\models \K\varphi$ if{f} $\langle W, \pi, R_K, R_A, w\rangle \models \varphi$ for any $w\in W$ such that $(s, w)\in R_K$,
\item $M\models \A\varphi$ if{f} $\langle W, \pi, R_K, R_A, w\rangle \models \varphi$ for any $w\in W$ such that $(s, w)\in R_A$.
\end{itemize}
Note that for any $w \in W$, $\pi(w)$ is an interpretation.
We say that a Kripke interpretation~$M$ is a {\em model} of a GK formula~$\varphi$ if $M$ satisfies $\varphi$, $M$ is a {\em model} of a GK theory $T$ if $M$ satisfies every GK formula in $T$.
In the following, given a Kripke interpretation $M$, we let
\begin{align*}
\K(M) &= \{\, \phi\mid \phi \textrm{ is a propositional formula and } M\models \K \phi\,\},\\
\A(M) &= \{\, \phi\mid \phi \textrm{ is a propositional formula and } M\models \A \phi\,\}.
\end{align*}
Notice that $\K(M)$ and $\A(M)$ are always closed under classical logical entailment -- they are propositional theories.

Given a GK formula $T$, a Kripke interpretation $M$ is a minimal model of~$T$ if $M$ is a model of~$T$ and
there does not exist another model $M_1$ of $T$ such that $\A(M_1) = \A(M)$ and $\K(M_1) \subsetneq \K(M)$.
We say that $M$ is a {\em GK model} of $T$ if $M$ is a minimal model of~$T$ and $\K(M) = \A(M)$.

In this paper, we consider only GK formulas that do not contain nested occurrences of modal operators.
Specifically, an~{\em $\A$-atom} is a formula of the form $\A\phi$ and a {\em $\K$-atom} is a formula of the form $\K\phi$, where $\phi$ is a propositional formula.
A GK formula is called a {\em pure GK formula} if it is formed from $\A$-atoms, $\K$-atoms and propositional connectives.
Similarly, a {\em pure GK theory} is a set of pure GK formulas.
Given a pure GK formula $F$, we denote
\begin{align*}
\Katom(F) & = \{\, \phi \mid \K \phi \text{ is a $\K$-atom occurring in $F$}\,\},\\
\Aatom(F) & = \{\, \phi \mid \A \phi \text{ is an $\A$-atom occurring in $F$}\,\}.
\end{align*}
For a pure GK theory $T$, we use
$\Katom(T) = \bigcup_{F\in T}\Katom(F)$ and $\Aatom(T) = \bigcup_{F\in T}\Aatom(F)$ to denote their modal atoms.

So far, the applications of the logic of GK only ever use pure GK formulas.
We now present some embeddings of well-known nonmonotonic knowledge representation languages into the logic of GK.

\myparagraph{Default logic}
A (propositional) default theory $\Delta=(W,D)$ (under extension semantics) is translated into pure GK formulas
in the following way:
(1) Translate each $\phi\in W$ to $\K \phi$;
(2) translate each $(\phi:\psi_1,\dots,\psi_n/\varphi)\in D$ to
\mbox{$\K \phi\land\neg \A\neg \psi_1\land\dots\land\neg \A\neg \psi_n\supset \K \varphi$}.
For the weak extension semantics, a default $(\phi:\psi_1,\dots,\psi_n/\varphi)\in D$ is translated to
\mbox{$\A \phi\land\neg \A\neg \psi_1\land\dots\land\neg \A\neg \psi_n\supset \K \varphi$}.

\myparagraph{Autoepistemic logic}
An $L$-sentence of autoepistemic logic that is in normal form~\citep{konolige88ontherelation}, that is, a disjunction of the form
\mbox{$\neg L \phi \lor L \psi_1 \lor \cdots \lor L \psi_n \lor \varphi$},
is (under expansion semantics) expressed as
\mbox{$\A \phi \land \neg \A \psi_1 \land \cdots \land \neg \A \psi_n \supset \K \varphi$}.
For strong expansion semantics, it becomes
\mbox{$\K \phi \land \neg \A \psi_1 \land \cdots \land \neg \A \psi_n \supset \K \varphi$}.

Notice that the translation of default and autoepistemic theories into the logic of GK is compatible with Konolige's translation from default logic into autoepistemic logic~\citep{konolige88ontherelation}.
Indeed, Konolige's translation perfectly aligns the weak extension semantics of default logic with expansion semantics for autoepistemic logic, and likewise for extension and strong expansion semantics~\citep{denecker03uniformsemantic}.

\myparagraph{Logic of universal causation}
The logic of universal causation is a nonmonotonic propositional modal logic with one modality $\C$~\citep{Turner}.
A formula of this logic is translated to the pure logic of GK by replacing every occurrence of $\C$ by $\K$, adding $\A$ before each atom which is not in the range of $\C$ in it, and adding $\A p\lor \A \neg p$ for each atom $p$. For example, if a UCL formula is $(p\land \neg q)\supset \C(p\land \neg q)$ and $\Atom = \{p, q\}$, then the corresponding pure GK formula is $\left((\A p\land \neg \A q) \supset \K(p\land \neg q)\right) \land (\A p\lor \A\neg p) \land (\A q\lor \A \neg q)$.

\myparagraph{Disjunctive logic programs}
A disjunctive LP rule
\[
p_1\lor\cdots\lor p_k\gets p_{k+1}, \ldots, p_l, \Not  p_{l+1},
\ldots, \Not p_{m},
\]
where $p$'s are atoms,
corresponds to the pure GK formula:
\[
\K p_{k+1}\land\cdots\land \K p_l\land  \neg \A p_{l+1}\land\cdots\land
\neg \A p_{m}\supset
\K p_1\lor\cdots\lor \K p_k
\]

\presection
\section{Main Result: From Pure GK to Disjunctive ASP}


Before presenting the translation, we introduce some notations.
Let $F$ be a pure GK formula, we use $\tr_p(F)$ to denote the propositional formula obtained from $F$ by replacing each occurrence of a $\K$-atom $\K \phi$ by $\katom\phi$ and each occurrence of an $\A$-atom $\A \psi$ by $\aatom\psi$, where $\katom\phi$ and $\aatom\psi$ are new atoms with respect to $\phi$ and $\psi$ respectively.
For a pure GK theory $T$, we define $\tr_p(T) = \bigwedge_{F\in T} \tr_p(F)$.
To illustrate these and the definitions that follow, we use a running example.
\begin{firstexample}{exm:gk}{Normal Reiter default}
  Consider the pure GK theory $\{ F \}$ with $F=\neg\A \neg p \supset \K p$ corresponding to the default $\tdefault{\top}{p}{p}$, and
  another pure GK theory $\{ F, G\}$ with $G = \K \neg p$ corresponding to the default $\tdefault{\top}{\top}{\neg p}$.
  Then $\tr_p(\{F\}) = \neg\aatom{\neg p} \supset \katom{p}$ and $\tr_p(\{F, G\}) = (\neg\aatom{\neg p} \supset \katom{p}) \land \katom{\neg p}$, where $\aatom{\neg p}$, $\katom{p}$, and $\katom{\neg p}$ are new atoms.
\end{firstexample}
Here we introduce a set of new atoms $\katom\phi$ and $\aatom\psi$ for each formula $\phi\in \Katom(T)$ and $\psi\in \Aatom(T)$.
Intuitively, the new atom $\katom\phi$ (resp.~$\aatom\psi$) will be used to encode containment of the formula $\phi$ in $\K(M)$ (resp.~$\A(M)$) of a GK model $M$ for $T$.

Given a propositional formula $\phi$ and an atom $a$, we use $\phi^a$ to denote the propositional formula obtained from $\phi$ by replacing each occurrence of an atom $p$ with a new atom $p^a$ with respect to $a$.
These formulas and new atoms will later be used in our main translation to perform the minimality check of the logic of GK's semantics.

We now stepwise work our way towards the main result.
We start out with a result that relates a pure GK theory to a propositional formula that will later reappear in our main translation.

\begin{proposition}\label{prop:1}
Let $T$ be a pure GK theory. A Kripke interpretation $M$ is a model of $T$ if and only if there exists a model $I^*$ of the propositional formula
$\formOne_T$ where
\begin{align*}
  \formOne_T & = \tr_p(T) \land \formOne_{\sound} \land \formOne^\K_{\witness} \land \formOne^\A_{\witness} \text{ with} \\
  \formOne_{\sound} &= \bigwedge_{\phi\in \Katom(T)} (\katom\phi \supset \phi^{\katom{}}) \land
  \bigwedge_{\phi\in \Aatom(T)} (\aatom\phi\supset \phi^{\aatom{}}) \\
  \formOne^\K_{\witness} &= \bigwedge_{\psi\in \Katom(T)} \left( \neg \katom\psi  \supset
    \formOne^\K_\psi
  \right) \\
  \formOne^\A_{\witness}&=\bigwedge_{\psi\in \Aatom(T)} \left( \neg \aatom\psi  \supset
    \formOne^\A_\psi
  \right) \\
  \formOne^\K_\psi &= \neg \psi^{\katom\psi} \land \bigwedge_{\phi\in \Katom(T)} (\katom\phi\supset \phi^{\katom\psi}) \\
  \formOne^\A_\psi &= \neg \psi^{\aatom\psi} \land \bigwedge_{\phi\in \Aatom(T)} (\aatom\phi\supset \phi^{\aatom\psi})
\end{align*}
such that
\begin{itemize}
\item $\K(M)\cap \Katom(T) = \{\phi \mid \phi \in \Katom(T),\, I^*\models \katom\phi\}$;
\item $\A(M)\cap \Aatom(T) = \{\phi \mid \phi \in \Aatom(T),\, I^*\models \aatom\phi\}$.
\end{itemize}
\end{proposition} 


The proposition examines the relationship between models of a pure GK theory and particular models of the propositional formula~$\formOne_T$. 
The first conjunct $\tr_p(T)$ of the formula $\formOne_T$ indicates that 
the $\katom{}$-atoms and $\aatom{}$-atoms in it can be interpreted in accordance with $\K(M)$ and $\A(M)$ such that $I^*\models \tr_p(T)$ iff $M$ is a model of $T$.
The soundness formula $\formOne_\sound$ achieves that the sets $\{\phi \mid \phi \in \Katom(T) \text{ and } I^*\models \katom\phi\}$ and $\{\phi \mid \phi \in \Aatom(T) \text{ and } I^*\models \aatom\phi\}$ are consistent.
The witness formulas $\formOne_{\witness}$ indicate that, if $I^*\models \neg \katom\psi$ for some $\psi\in \Katom(T)$ (resp.~$\psi\in \Aatom(T)$) then there exists a model $I'$ of $\K(M)$ (resp.~$\A(M)$) such that $I'\models \neg \psi$, where $I'$ is explicitly indicated by newly introduced $p^{\katom\psi}$ (resp.~$p^{\aatom\psi}$) atoms.
So intuitively, if a formula is not known (or not assumed), then there must be a witness for that.
This condition is necessary:
for instance, the set $\{ \katom{p}, \katom{q}, \neg \katom{p\land q}\}$ satisfies the formula $(\katom{p\land q}\supset \katom{p}) \land (\katom{p\land q} \supset \katom{q})$, however, since $\K(M)$ is a theory there does not exist a Kripke interpretation $M$ such that $p\in \K(M)$, $q\in \K(M)$ and $p\land q \notin \K(M)$.

\begin{cexample}{exm:gk}
  Formula $\formOne_{\{F\}}$ 
  is given by:
  \begin{align*}
    \tr_p(\{F\}) &= \neg\aatom{\neg p} \supset \katom{p} \\
    \formOne_{\sound}(\{F\}) &= ( \katom{p} \limplies {p}^{\katom{}} ) \land ( \aatom{\neg p} \limplies {\neg p}^{\aatom{}} ) \\
    \formOne^\K_{\witness}(\{F\}) &= \neg\katom{p} \limplies ( \neg {p}^{\katom{p}} \land ( \katom{p} \limplies {p}^{\katom{p}} ) ) \\
    \formOne^\A_{\witness}(\{F\}) &= \neg\aatom{\neg p} \limplies ( \neg\neg {p}^{\aatom{\neg p}} \land ( \aatom{\neg p} \limplies {\neg p}^{\aatom{\neg p}} ) )
  \end{align*}
  Formula $\formOne_{\{F, G\}}$ is given by:
  \begin{align*}
    tr_p(\{F, G\}) &= (\neg\aatom{\neg p} \supset \katom{p}) \land \katom{\neg p}\\
    \formOne_{\sound}(\{F, G\}) &= \formOne_{\sound}(\{F\}) \land ( \katom{\neg p} \limplies \neg {p}^{\katom{}} ) \\
    \formOne^\K_{\witness}(\{F, G\}) &= (\neg\katom{p} \limplies \formOne^\K_{p}) \land (\neg \katom{\neg p} \limplies \formOne^\K_{\neg p})\\
    \formOne^\A_{\witness}(\{F, G\}) &= \formOne^\A_{\witness}(\{F\})\\
    \formOne^\K_{p} &= \neg {p}^{\katom{p}} \land ( \katom{p} \limplies {p}^{\katom{p}}) \land (\katom{\neg p} \limplies \neg {p}^{\katom{p}})\\
    \formOne^\K_{\neg p} &= \neg \neg {p}^{\katom{\neg p}} \land ( \katom{p} \limplies {p}^{\katom{\neg p}}) \land (\katom{\neg p} \limplies \neg {p}^{\katom{\neg p}})
  \end{align*}
  where $p^{\katom{}}$, $p^{\aatom{}}$, $p^{\katom{p}}$, $p^{\aatom{\neg p}}$, and $p^{\katom{\neg p}}$ are new atoms.
  Note that formula $\formOne_{\sound}(\{F, G\})$ prevents a model that satisfies both $\katom{p}$ and $\katom{\neg p}$.
\end{cexample}

While Proposition~\ref{prop:1} aligns Krikpe models and propositional models of the translation, there is yet no mention of GK's typical minimization step.
This is the task of the next result, which extends the above relationship to GK models.

\begin{proposition}\label{prop}
Let $T$ be a pure GK theory. A Kripke interpretation $M$ is a GK model of $T$ if and only if there exists a model $I^*$ of the propositional formula $\formOne_T$ 
such that
\begin{itemize}
\item \mbox{$\K(M) = \A(M) = \Th\left(\, \{\phi \mid \phi \in \Atom_\K(T), I^*\models \katom\phi\}\,\right)$};
\item for each $\psi\in \Atom_\A(T)$,
 \[ I^*\models \aatom\psi \text{ iff } \psi \in \Th(\{\phi\mid \phi\in \Atom_\K(T) \text{ and } I^*\models \katom\phi\}) \]
\item there does not exist another model $I^{*\prime}$ such that
  \begin{align*}
    \hspace*{-4mm} I^{*\prime} \cap \{ \aatom\phi \mid \phi\in \Atom_\A(T)\} &= I^* \cap \{\aatom\phi \mid \phi\in \Atom_\A(T)\}, \\
    \hspace*{-4mm} I^{*\prime} \cap \{ \katom\phi \mid \phi\in \Atom_\K(T)\} &\subsetneq I^*\cap \{ \katom\phi \mid \phi\in \Atom_\K(T)\}.
  \end{align*}
\end{itemize}
\end{proposition}

\begin{cexample}{exm:gk}
  Clearly the intended reading of our running example $\{F\}$ is that there is no reason to assume that $p$ is false, and the default lets us conclude that we know $p$.
  This is testified by the partial interpretation
  $I^* = \{ \neg\aatom{\neg p}, \katom{p}, {p}^{\katom{}}, {p}^{\aatom{\neg p}} \}$
  (the remaining atoms are not relevant).
  It is easy to see that $I^*$ is a model for $\formOne_{\{F\}}$ 
  and there is no model ${I^*}'$ with the properties above.
  Now \mbox{$\katom{p}\in I^*$} shows that $p$ is known in the corresponding GK model. 
  
  Similarly, $G$ provides a reason to assume that $p$ is false and $\{F, G\}$ concludes that we know $\neg p$. Consider the partial interpretation $I^* = \{ \aatom{\neg p}, \neg \katom{p}, \katom{\neg p}, \neg {p}^{\katom{}}, \neg {p}^{\aatom{}}, \neg {p}^{\katom{p}}\}$, it specifies a model for $\formOne_{\{F, G\}}$ and there is no model $I^{*\prime}$ with the properties above. In particular, $\katom{\neg p}\in I^*$ shows that $\neg p$ is known in the corresponding GK model.
\end{cexample}

In Proposition~\ref{prop}, we only need to consider a Kripke interpretation $M$ such that $\A(M)\cup \K(M)$ is consistent.
This means that formula $\formOne_T$ 
can be modified to $\formTwo_T$ where
\begin{align*}
  \formTwo_T &= \tr_p(T) \land \formTwo_{\sound} \land \formTwo^\K_{\witness} \land \formTwo^\A_{\witness} \text{ with} \\
  \formTwo_{\sound} &= \bigwedge_{\phi\in \Katom(T)} (\katom\phi \supset \phi) \land
  \bigwedge_{\phi\in \Aatom(T)} (\aatom\phi \supset \phi) \\
  \formTwo^\K_{\witness} &= \bigwedge_{\psi\in \Katom(T)} \Bigg( \neg \katom\psi \supset \formTwo^\K_\psi \Bigg) \\
  \formTwo^\A_{\witness} &= \bigwedge_{\psi\in \Aatom(T)} \Bigg( \neg \aatom\psi \supset \formTwo^\A_\psi \Bigg) \\
  \formTwo^\K_\psi &= \neg \psi^{\katom\psi} \land \bigwedge_{\phi\in \Katom(T)} (\katom\phi\supset \phi^{\katom\psi})\ \land \\
  &\hspace*{3cm}\bigwedge_{\phi\in \Aatom(T)} (\aatom\phi\supset \phi^{\katom\psi}) \\
  \formTwo^\A_\psi &= \neg \psi^{\aatom\psi} \land \bigwedge_{\phi\in \Katom(T)} (\katom\phi\supset \phi^{\aatom\psi})\ \land \\
  &\hspace*{3cm}\bigwedge_{\phi\in \Aatom(T)} (\aatom\phi\supset \phi^{\aatom\psi})
\end{align*}

So the soundness formula $\formTwo_{\sound}$ actually becomes {easier}, since soundness of knowledge and assumptions is enforced for one and the same vocabulary (the one from the original theory).
The witness formulas become somewhat more complicated, as the witnesses have to respect both the knowledge as well as the assumptions of the theory.
This is best explained by consulting our running example again.

\begin{cexample}{exm:gk}
  While $F$'s propositionalization $\tr_p(\{F\})$ stays the same, the soundness and witness formulas change in the step from formula $\formOne_{\{F\}}$ 
  to formula $\formTwo_{\{F\}}$. 
  We only show the first conjunct of the witness formula $\formTwo_{\witness}$, which is given by
  \begin{gather*}
    \neg \katom{p} \limplies \left( \neg p^{\katom{p}} \land
      \left( \katom{p} \limplies p^{\katom{p}} \right) \land
      \left( \aatom{\neg p} \limplies \neg p^{\katom{p}} \right) \right)
  \end{gather*}
  Intuitively, the formula expresses that whenever $p$ is not known, then there must be a witness, that is, an interpretation where $p$ is false.
  Since the witnessing interpretations could in principle be distinct for each $\K$-atom, they have to be indexed by the respective $\K$-atom they refer to, as in $p^{\katom{p}}$.
  Of course, the witnesses have to obey all that is known and assumed, which is guaranteed in the last two conjuncts.
\end{cexample}

Using this new formula, the result of Proposition~\ref{prop} can be restated.

\begin{proposition}\label{prop:2}
Let $T$ be a pure GK theory.
A Kripke interpretation $M$ is a GK model of $T$ if and only if there exists a model $I^*$ of the propositional formula~$\formTwo_T$ 
such that
\begin{itemize}
\item \mbox{$\K(M) = \A(M) = \Th\left(\{\phi \mid \phi \in \Atom_\K(T), I^*\models \katom\phi\}\right)$};
\item for each $\psi\in \Atom_\A(T)$, we have that $I^*\models \aatom\psi$ implies
\[
\psi \in \Th(\{\phi\mid \phi\in \Atom_\K(T) \text{ and } I^*\models \katom\phi\})
\]
\item there does not exist another model $I^{*\prime}$ of~$\formOne_T$ 
such that
\begin{align*}
  \hspace*{-3mm} I^{*\prime} \cap \{ \aatom\phi \mid \phi\in \Atom_\A(T)\} &= I^* \cap \{\aatom\phi \mid \phi\in \Atom_\A(T)\} \\
  \hspace*{-3mm} I^{*\prime} \cap \{ \katom\phi \mid \phi\in \Atom_\K(T)\} &\subsetneq I^*\cap \{ \katom\phi \mid \phi\in \Atom_\K(T)\}
\end{align*}
\end{itemize}
\end{proposition}

We are now ready for our main result, translating a pure GK theory to a disjunctive logic program.
First, we introduce some notations.
Let $T$ be a pure GK theory, we use $\trne(T)$ to denote the nested expression obtained from~$\formTwo_T$ 
by first converting it to negation normal form\footnote{A propositional formula is in Negation Normal Form (NNF) if negation occurs only immediately above atoms, and $\{ \bot, \top, \neg, \land, \lor\}$ are the only allowed connectives.}, then
replacing ``$\land$'' by ``$,$'' and ``$\lor$'' by ``$;$''.
A propositional formula $\phi$ can be equivalently translated to conjunctive normal form (involving at most linear blowup) 
\begin{multline*}
  (p_1 \lor \cdots \lor p_t \lor \neg p_{t+1} \lor \cdots \lor\neg p_m) \land \ldots \\
  \land (q_1 \lor \cdots \lor q_k \lor \neg q_{k+1} \lor \cdots \lor \neg q_n)
\end{multline*}
where $p$'s and $q$'s are atoms;
we use $\trc(\phi)$ to denote the set of rules
\begin{gather*}
p_1; \ldots ; p_t \gets p_{t+1}, \ldots, p_m
\quad\ldots\quad
q_1; \ldots; q_k \gets q_{k+1}, \ldots, q_n
\end{gather*}
We use $\widehat{\phi}$ to denote the propositional formula obtained from $\phi$ by replacing each occurrence of an atom $p$ by a new atom $\hat{p}$.

We use $T^*$ to denote the propositional formula obtained from the formula~$\formOne_T$ 
by replacing each occurrence of an atom $p$ (except atoms in $\{\aatom\phi\mid \phi\in\Atom_\A(T)\}$) by a new atom $p^*$. Intuitively, each atom that is not an $\aatom{}$-atom is replaced by a new atom.

Notice that $\trne(T)$ is obtained from~$\formTwo_T$ 
while $T^*$ is obtained from~$\formOne_T$. 
Intuitively, by Proposition~\ref{prop:2}, $\trne(T)$ is used to restrict interpretations for introduced $\katom{}$-atoms and $\aatom{}$-atoms so that these interpretations serve as candidates for GK models, and by Proposition~\ref{prop:1}, $T^*$ constructs possible models of the GK theory which are later used to test whether these models prevent the candidate to be a GK model.

\begin{figure*}[t]
  \centering
  \begin{align*}
    \linenr&& \bot &\gets \Not \trne(T) \\
    \linenr&& p';\neg p' &\gets \top &\textrm{\em(for each atom $p'$ occurring in $\trne(T)$)}\\
    \linenr&& u; A &\gets B &\textrm{\em(for each rule $A\gets B$ in $\trc(T^*)$)} \\
    \linenr&& u; c_{\phi_1}; \cdots; c_{\phi_m}&\gets \top &\textrm{\em ($\{{\phi_1}, \ldots, {\phi_m}\}=\Katom(T)$)} \\
    \linenr&& u &\gets c_\phi, \Not \katom\phi &\textrm{\em(for each $\phi\in \Atom_\K(T)$)} \\
    \linenr&& u &\gets k^*_\phi, \Not \katom\phi &\textrm{\em(for each $\phi\in \Atom_\K(T)$)} \\
    \linenr&& u &\gets c_{\phi}, k^*_\phi, \Not \neg \katom\phi &\textrm{\em(for each $\phi\in \Atom_\K(T)$)} \\
    \linenr&& u; c_\phi; k^*_\phi &\gets \Not \neg \katom\phi &\textrm{\em(for each $\phi\in \Atom_\K(T)$)} \\
    \linenr&& p^{*} &\gets u &\textrm{\em(for each new atom $p^{*}$ occurring in $\trc(T^*)$)} \\
    \linenr&& c_\phi &\gets u &\textrm{\em(for each $\phi\in \Atom_\K(T)$)} \\
    \linenr&& \bot &\gets \Not u\\
    \linenr&& v; A &\gets B &\textrm{\em(for each rule $A\gets B$ in} \\
           &&      &        &\trc\left(\bigwedge_{\phi\in \Atom_\K(T)}(\katom\phi \supset \widehat{\phi})\land \neg \bigwedge_{\phi\in \Atom_\A(T)}(\aatom\phi \supset\widehat{\phi})\right) \textrm{\em)} \\
    \linenr&& \hat{p} &\gets v &\textrm{\em(for each atom $\hat{p}$  except $\katom{}$-atoms and $\aatom{}$-atoms occurring in} \\
           &&         &        &\trc\left(\bigwedge_{\phi\in \Atom_\K(T)}(\katom\phi \supset \widehat{\phi})\land \neg \bigwedge_{\phi\in \Atom_\A(T)}(\aatom\phi \supset\widehat{\phi})\right) \textrm{\em)} \\
    \linenr&& \bot &\gets \Not v
  \end{align*}
  \caption{Translation from pure GK theory $T$ to disjunctive logic program $\trlp(T)$ used in Theorem~\ref{them}, where $u$, $v$, and $c_\phi$ (for each $\phi\in \Katom(T)$) are new atoms.}
  \label{fig:translation}
\end{figure*}

Inspired by the linear translation from parallel circumscription into disjunctive logic programs by \citet{janhunen2004capturing}, we have the following theorem.

\begin{theorem}\label{them}
Let $T$ be a pure GK theory. A Kripke interpretation $M$ is a GK model of $T$ if and only if there exists an answer set $S$ of the logic program $\trlp(T)$ in Figure~\ref{fig:translation}
with \mbox{$\K(M)= \A(M) = \Th( \{\phi\mid \phi\in \Atom_\K(T) \text{ and } \katom\phi\in S\})$}.
\end{theorem}


The intuition behind the construction is as follows:
\begin{itemize}
\item (1) and (2) in $\trlp(T)$: $I^*$ is a model of the formula~$\formTwo_T$. 
\item (3--8): if there exists a model $I^{*\prime}$ of the formula~$\formOne_T$ 
with
\begin{align*}
  \hspace*{-3mm} I^*\cap \{\aatom\phi\mid \phi\in\Atom_\A(T)\} &= I^{*\prime} \cap \{\aatom\phi\mid \phi\in\Atom_\A(T)\} \\
  \hspace*{-3mm} I^{*\prime}\cap \{\katom\phi\mid \phi\in\Atom_\K(T)\} &\subsetneq I^*\cap\{\katom\phi\mid \phi\in \Atom_\K(T)\},
\end{align*}
then there exists a set $S^*$ constructed from new atoms in $\trc(T^*)$ (which is a copy of the formula~$\formOne_T$ 
with same $\aatom\phi$ for each $\phi\in \Atom_\A(T)$) and $c_\phi$ for some $\phi\in \Atom_\K(T)$ such that $S^*$ satisfies rules (3) to (8) and $u\notin S^*$.
\item (9) and (10): if there is such a set $S^*$ then it is the least set containing $u$, all $p^*$'s and $c$-atoms.
\item (11): such a set $S^*$ should not exist.
  (See item~3 in Proposition~\ref{prop:2}.)
\item (12) and (13): if there exists a model of the formula $\bigwedge_{\phi\in\Atom_\K(T)} (\katom\phi \supset \widehat{\phi}) \land \neg \bigwedge_{\phi\in \Atom_\A(T)}(\aatom\phi \supset \widehat{\phi})$, then $v$ should not occur in the minimal model of the program.
\item (14): $\bigwedge_{\phi\in\Atom_\K(T)} (\katom\phi \supset \widehat{\phi}) \land \neg \bigwedge_{\phi\in \Atom_\A(T)}(\aatom\phi \supset \widehat{\phi})$ should not be consistent.
  (This is necessary by item~2 in Proposition~\ref{prop:2}.)
\end{itemize}

Given a model $S$ of the logic program $\trlp(T)$, the new atom $u$ is used to indicate that the model $I^*$ of $\formTwo_T$ w.r.t. $S$ (specified by (1) and (2)) satisfies item 3 in Proposition~\ref{prop:2}. Specifically, if $I^*$ does not satisfy item 3, then there exists a subset $S^*$ of $p^*$'s and $c$-atoms that satisfies (3) to (8). If in addition $u\notin S^*$, then there exists a subset of $S$ that satisfies all rules in $\trlp(T)$ except (11), thus $S$ cannot be an answer set of $\trlp(T)$.
Similarly, $v$ is used to indicate that $I^*$ satisfies item 2 in Proposition~\ref{prop:2}. Specifically, if $I^*$ does not satisfy item 2, then the propositional formula $\bigwedge_{\phi\in\Atom_\K(T)} (\katom\phi \supset \widehat{\phi}) \land \neg \bigwedge_{\phi\in \Atom_\A(T)}(\aatom\phi \supset \widehat{\phi})$ is satisfiable, thus there exists a subset $\widehat{S}$ of $\hat{p}$'s that satisfies (12).
If in addition $v\notin \widehat{S}$, then there exists a subset of $S$ that satisfies all rules in $\trlp(T)$ except (14), thus $S$ cannot be an answer set of $\trlp(T)$.

\begin{cexample}{exm:gk}
  For our running example theory $\{F\}$ with \mbox{$F=\neg\A\neg p\supset \K p$}, we find that the logic program translation $\trlp(\{F\})$ has a single answer set $S$ with $\katom{p}\in S$
  Thus by Theorem~\ref{them} we can conclude that the GK theory $\{F\}$ has a single GK model $M$ in which $\K(M)=\Th(\{p\})$.
  Likewise, the logic program $\trlp(\{F, G\})$ has a single answer set $S'$ with $\katom{\neg p} \in S'$, whence $\{F, G\}$ has a single GK model $M'$ in which $\K(M') = \Th(\{\neg p\})$.
\end{cexample}

\myparagraph{Computational complexity}
We have seen in the preliminaries section that disjunctive logic programs can be modularly and equivalently translated into pure formulas of the logic of GK.
Conversely, Theorem~\ref{them} shows that pure GK formulas can be equivalently translated into disjunctive logic programs.
Eiter and Gottlob showed that the problem of deciding whether a disjunctive logic program has an answer set is $\Sigma^P_2$-complete~\citep{eiter-gottlob95dasp-complexity}.
In combination, these results yield the following straightforward complexity result for the satisfiability of pure GK.

\begin{proposition}
  \label{p:complexity}
  Let $T$ be a pure GK theory.
  The problem of deciding whether $T$ has a GK model is $\Sigma^P_2$-complete.
\end{proposition}
We remark that the hardness of disjunctive logic programs stems from so-called head cycles (at least two atoms that mutually depend on each other and occur jointly in some rule head).
It is straightforwardly checked that our encoding creates such head cycles, for example the head of rule (8) contains the cycle induced by rules (7) and (10).

\presection
\presection
\presection
\section{Implementation}

We have implemented the translation of Theorem~\ref{them} into a working prototype \gkdlp.
The program is written in Prolog and uses the disjunctive ASP solver claspD-2~\citep{gebser13claspD2}, which was ranked first place in the 2013 ASP competition.\footnote{\url{http://www.mat.unical.it/ianni/storage/aspcomp-2013-lpnmrtalk.pdf}}

Our prototype is the first implementation of the (pure) logic of GK to date.
The restriction to pure formulas seems harmless since all known applications of the logic of GK use only pure formulas.
We remark that \gkdlp{} implements default and autoepistemic logics such that input and target language are of the same complexity. 

\myparagraph{Evaluation}
To have a scalable problem domain and inspired by dl2asp~\citep{chen10dl2asp}, we chose the fair division problem~\citep{bouveret08efficiency} for experimental evaluation.
An instance of the fair division problem consists of a set of agents, a set of goods, and for each agent a set of constraints that intuitively express which sets of goods the agent is willing to accept.
A solution is then an assignment of goods to agents that is a partition of all goods and satisfies all agents' constraints.
\citet{bouveret08efficiency} showed that the problem is $\Sigma^P_2$-complete, and can be naturally encoded in default logic.

We created random instances of the fair division problem with increasing numbers of agents and goods.
We then applied the translation of \citep{bouveret08efficiency}, furthermore the translation from default logic into the logic of GK, then invoked \gkdlp{} to produce logic programs and finally used gringo~3.0.3 and claspD version~2 (revision~6814) to compute all answer sets of these programs, thus all extensions of the original default theory corresponding to all solutions of the problem instance.
The experiments were conducted on a Lenovo laptop with an Intel Core~i3 processor with 4 cores and 4GB of RAM running Ubuntu~12.04.
We recorded the size of the default theory, the size of the translated logic program, the translation time and the solving time, as well as the number of solutions obtained.
We started out with 2 agents and 2 goods, and stepwise increased these numbers towards 6.
For each combination in \mbox{$(a,g)\in \set{2,\ldots,6}\times\set{2,\ldots,6}$}, we tested 20 randomly generated instances.
Random generation here means that we create agents' preferences by iteratively drawing random subsets of goods to add to an agent's acceptable subsets with probability $P$, where $P$ is initialized with $1$ and discounted by the factor $\frac{g-1}{g}$ for each subset that has been drawn.

In accordance with our theoretical predictions, we observed that the increase in size from GK formula to logic program is indeed polynomial (albeit with a low exponent).
The plot on the right (Figure~\ref{fig:time}) shows the solving time in relation to the size of the default theory, where the time axis is logarithmic.
We can see that the runtime behavior of \gkdlp{} is satisfactory.
We acknowledge however that the runtimes we measured are not competitive with those reported by \citet{chen10dl2asp} for dl2asp.
However, a direct comparison of the two systems is problematic for a number of reasons.
First of all, the system dl2asp is not publicly available to the best of our knowledge.
Furthermore, \citet{chen10dl2asp} do not describe how they create random instances of the fair division problem, so we cannot compare the runtimes they report and the ones we measured.
Finally, dl2asp is especially engineered for default logic, and it is not clear how their approach can be generalized to other languages, for example Turner's logic of universal causation. 
In general, the approaches to translation that are followed by dl2asp and \gkdlp{} are completely different:
dl2asp translates a $\Sigma^P_2$-complete problem to an $\mathsf{NP}$-complete problem using a translation in $\Delta^P_2$.
Our system \gkdlp{} translates a $\Sigma^P_2$-complete problem into another $\Sigma^P_2$-complete problem using a translation that can be computed in polynomial time.

\def\xscale{0.66}
\def\yscale{0.66}
\def\figurespace{-3mm}


\begin{figure}[h]
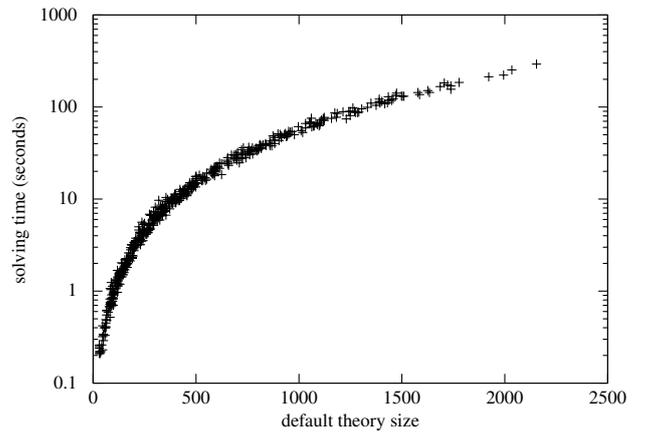

  \centering
  {\smaller[2]
    \iffinal{\include{fd-6-6-20-time.csv}}\fi
  }
  \vspace*{\figurespace}
  \caption{Solving time (log scale) with respect to default theory size.}
  \vspace*{\figurespace}
  \label{fig:time}
\end{figure}

\myparagraph{Applications}
We see immediate applicability of the translation of the present paper to several areas.
\citet{reiter87diagnosis} provided a theory of diagnosis from first principles, and showed how default logic can be used as an implementation device.
\citet{cadoli94defaultlogic} proposed to use default logic as an expressive query language on top of relational databases, and gave an example of achieving strategic behavior in an economic setting.
In reasoning about actions, \citet{thielscher96causalityand} used default logic to solve the qualification problem of dealing with unexpected action failures.
\citet{martin01addressing} later provided an implementation of that approach where extensions are enumerated in Prolog.
Recently, \citet{baumann10statedefaults} introduced a method for default reasoning in action theories, that is, an approach to the question what normally holds in a dynamic domain.
Our translation yields an implementation of their approach, something that they stated as future work and later achieved to a limited extent (for a restricted sublanguage of their framework~\citep{strass12draculasp}).
In a similar vein, \citet{pagnucco13implementing} looked at belief change in the situation calculus and proposed an implementation based on default logic with preferences~\citep{brewka94addingpriorities,delgrande-schaub00expressingpreferences}.

\myparagraph{Related work}
The translation presented in this paper is a generalization of the one presented for Turner's logic of universal causation by \citet{ji-lin13turners}.
We chose the logic of GK as general nonmonotonic language, we could also have chosen the logic of minimal belief and negation as failure~\citep{lifschitz94mbnf}, the logic of here-and-there~\citep{heyting30ht} or the nonmonotonic modal logic S4F~\citep{schwarz-truszczynski94minimal-knowledge}.
In terms of implementations, there are few approaches that treat as broad a range of propositional nonmonotonic knowledge representation languages as \gkdlp.
Notable exceptions are the works of \citet{junker-konolige90computing}, who implemented both autoepistemic and default logics by translating them to truth maintenance systems;
\citet{niemela95decision}, who provides a decision procedure for autoepistemic logic which also incorporates extension semantics for default logics;
and \citet{rosati99reasoning}, who provides algorithms for \citeauthor{lifschitz94mbnf}' logic of minimal belief and negation as failure~[\citeyear{lifschitz94mbnf}].
Other approaches are restricted to specific languages, where default logic seems to be most popular.
The recent system dl2asp~\citep{chen10dl2asp} translates default theories to normal (non-disjunctive) logic programs; the translation figures out all implication relations between formulas occurring in the default theory, just as \citet{junker-konolige90computing} did.
The authors of dl2asp~\citep{chen10dl2asp} already observed that default logic and disjunctive logic programs are of the same complexity;
they even stated the search for a polynomial translation from the former to the latter (that we achieved in this paper) as future work.
Gadel~\citep{nicolas00gadel} uses a genetic algorithm to compute extensions of a default theory;
likewise the system DeReS~\citep{cholewinski99computing} is not translation-based but directly searches for extensions;
similarly the XRay system~\citep{schaub97xray} implements local query-answering in default logics.
\citet{risch94tableaux} describe a tableaux-based algorithm for computing all extensions of general default theories, but do not report runtimes for their Prolog-based implementation.
For autoepistemic logic, \citet{marek91computing} investigate sceptical reasoning with respect to Moore's expansion semantics.

\presection
\section{Discussion}

We have presented the first translation of pure formulas of the logic of GK to disjunctive answer set programming.
Among other things, this directly leads to implementations of Turner's logic of universal causation as well as implementations of default and autoepistemic logics under different semantics.
We have prototypically implemented the translation and experimentally analysed its performance, which we found to be satisfactory given the system's generality.

In the future, we plan to integrate further nonmonotonic reasoning formalisms.
This is more or less straightforward due to the generality of this work:
to implement a language, it suffices to provide a translation into pure formulas of GK, then Theorem~\ref{them} of this paper does the rest.
Particular formalism we want to look at are
default logics with preferences~\citep{brewka94addingpriorities,delgrande-schaub00expressingpreferences} and
the logic of only-knowing~\citep{lakemeyer05onlyknowing}.
It also seems worthwhile to check whether our translation can be adapted to the nonmonotonic modal logic S4F~\citep{schwarz-truszczynski94minimal-knowledge,truszczynski07s4f}, that has only one modality instead of two.
We finally plan to study the approaches mentioned as applications in the previous section to try out our translation and implementation on agent-oriented AI problems.

{ \smaller

}

\appendix

\section*{Appendix}

\begin{longproof}
\textbf{ of Proposition~\ref{prop:1}:}

$\Rightarrow$: Let $M$ be a model of $T$, $I_1\subseteq \Lit$ a model of $\K(M)$, and $I_2\subseteq \Lit$ a model of $\A(M)$. Clearly, for each $\phi\in \Katom(T)$, if $\phi\in \K(M)$ then $I_1\models \phi$; if $\phi\notin \K(M)$ then there exists a model $I'$ of $\K(M)$ such that $I'\models \neg \phi$. Same results are established for each $\phi\in \Aatom(T)$.

Then, we can create an interpretation $I^*$ such that
\begin{multline*}
I^* = \{ l^{\katom{}} \mid l\in I_1\} \cup \{ l^{\aatom{}} \mid l\in I_2\}\\
 \cup\{ \katom\phi \mid \phi\in \Katom(T)\cap \K(M)\} \\  \cup\{ \aatom\phi\mid \phi\in \Aatom(T)\cap \A(M)\}\\
  \cup\{ \neg \katom\phi\mid \phi\in \Katom(T)\text{ and } \phi\notin \K(M)\}
\\ \cup\{\neg \aatom\phi\mid \phi\in \Aatom(T)\text{ and } \phi\notin \A(M)\} \\
 \cup \bigcup_{\psi\in \Katom(T)\atop \psi\in \K(M)} \{ l^{\katom\psi} \mid l\in I_1\} \cup \bigcup_{\psi\in \Aatom(T)\atop \psi\in \A(M)} \{ l^{\aatom\psi}\mid l\in I_2\} \\
\cup \bigcup_{\psi\in \Katom(T) \atop \psi\notin \K(M)} \left\{ l^{\katom\psi} \mid l\in I' \text{, $I'$ is a model of $\K(M) \cup \{\neg \psi\}$}\right\} \\
\cup \bigcup_{\psi\in \Aatom(T) \atop \psi\notin \A(M)} \left\{ l^{\aatom\psi} \mid l\in I' \text{, $I'$ is a model of $\A(M)\cup \{\neg \psi\}$}\right\}.
\end{multline*}
It is easy to verify that $I^*$ is a model of $\formOne_T$ and
\begin{itemize}
\item $\K(M)\cap \Katom(T) = \{\phi \mid \phi \in \Katom(T),\, I^*\models \katom\phi\}$;
\item $\A(M)\cap \Aatom(T) = \{\phi \mid \phi \in \Aatom(T),\, I^*\models \aatom\phi\}$.
\end{itemize}

$\Leftarrow$: Let $I^*$ be a model of $\formOne_T$. We can create a Kripke interpretation $M$ such that
\begin{itemize}
  \item $\K(M) = \Th\left(\, \{\phi \mid \phi \in \Katom(T) \text{ and } I^*\models \katom\phi\}\,\right)$;
  \item $\A(M) = \Th\left(\, \{\phi \mid \phi \in \Aatom(T) \text{ and } I^*\models \aatom\phi\}\,\right)$.
\end{itemize}

Note that, $\{ l\in \Lit\mid I^*\models l^{\katom{}}\}$ is a model of $\K(M)$ and $\{l\in \Lit \mid I^*\models l^{\aatom{}}\}$ is a model of $\A(M)$, then both $\K(M)$ and $\A(M)$ are consistent.

For each $\phi\in \Katom(T)$, if $I^*\models \katom\phi$ then $\phi\in \K(M)$; if $I^*\models \neg \katom\phi$ then there exists a model $I' = \{ l\in \Lit\mid I^*\models l^{\katom\phi}\}$ such that $I'$ is a model of $\K(M)$ and $I'\models \neg \phi$, thus $\phi\notin \K(M)$. So $I^*\models \katom\phi$ iff $\phi\in \K(M)$. The same result is established for each $\phi\in \Aatom(T)$.
Note that, $I^* \models \tr_p(T)$ then $M$ is a model of $T$.
\hfill
\end{longproof}

\begin{longproof}
\textbf{ of Proposition~\ref{prop}:}

$\Rightarrow$: Let $M$ be a GK model of $T$. From the proof of Proposition~\ref{prop:1}, we can create a model $I^*$ of $\formOne_T$.
Now we want to prove that $I^*$ satisfies all conditions in the proposition.

From Theorem~3.5 in~\citep{gk}, $\K(M) = \Th(\{\phi\mid \phi\in \Atom_\K(T)\cap\K(M)\})$, then $\K(M) =\A(M) = \Th(\{\phi\mid \phi\in \Atom_\K(T) \text{ and } I^* \models \katom\phi\})$.

Assume that there exists another model $I^{*\prime}$ of~$\formOne_T$ with
\begin{align*}
  I^{*\prime} \cap \{ \aatom\phi \mid \phi\in \Atom_\A(T)\} &= I^* \cap \{\aatom\phi \mid \phi\in \Atom_\A(T)\} \\
  I^{*\prime} \cap \{ \katom\phi \mid \phi\in \Atom_\K(T)\} &\subsetneq I^*\cap \{ \katom\phi \mid \phi\in \Atom_\K(T)\}
\end{align*}
Then, from Proposition~\ref{prop:1}, there exists a Kripke interpretation $M'$ such that $\K(M') = \Th(\{ \phi\mid \phi\in \Atom_\K(T) \text{ and } I^{*\prime}\models \katom\phi\})$, $\A(M') = \A(M)$, and $M'$ is a model of $T$. Note that, for each $\phi\in \Katom(T)$, $I^{*\prime} \models \neg \katom\phi$ implies $\K(M')\not\models \phi$, then $\K(M')\subsetneq \K(M)$.
From the definition of GK models, there does not exist such a model $M'$, which conflicts to the assumption, then there does not exist such a model $I^{*\prime}$.

From the construction of $I^*$, for each $\psi\in \Aatom(T)$, $I^*\models \aatom\psi$ iff $\psi\in \A(M)$.
Note that, $\K(M) =\A(M) = \Th(\{\phi\mid \phi\in \Atom_\K(T) \text{ and } I^* \models \katom\phi\})$, then $I^*\models \aatom\psi$ iff $\psi\in \Th(\{\phi\mid \phi\in \Atom_\K(T) \text{ and } I^* \models \katom\phi\})$.

So $I^*$ is a model of~$\formOne_T$ which satisfies all conditions in the proposition.

$\Leftarrow$: Let $I^*$ be a model of~$\formOne_T$ which satisfies corresponding conditions in the proposition. We can create a Kripke interpretation $M$ such that $\K(M) = \A(M) = \Th( \{\phi \mid \phi\in \Atom_\K(T) \text{ and } I^* \models \katom\phi\})$.

From the third condition in the proposition, $I^*\models \aatom\phi$ iff $\phi\in \K(M)$ for each $\phi\in\Aatom(T)$. Then $\A(M)\cap \Aatom(T) = \{ \phi\mid \phi\in \Aatom(T) \text{ and } I^*\models \aatom\phi\}$.
From the proof of Proposition~\ref{prop:1}, $M$ is a model of $T$ and $I^*\models \katom\phi$ (resp.~$I^*\models \aatom\phi$) iff $\phi\in \K(M)$ for each $\phi\in \Katom(T)$ (resp.~$\phi\in \Aatom(T)$). Now we want to prove that $M$ is a GK model of $T$.

Assume that there exists another model $M'$ of $T$ such that $\A(M') = \A(M)$ and $\K(M')\subsetneq \K(M)$. Note that $\K(M) = \Th( \{\phi \mid \phi\in \Atom_\K(T) \text{ and } I^* \models \katom\phi\})$, then $\K(M')\cap \Katom(T) \subsetneq \K(M)\cap \Katom(T)$.

Let $I = I^*\cap \{l^{\katom{}}\mid l\in \Lit\}$, clearly, $I$ is a model of $\K(M)$, $\A(M)$, and $\K(M')$. We can construct another model $I^{*\prime}$ of $\formOne_T$ as
\begin{multline*}
I^{*\prime} = \{ l^{\katom{}} \mid l\in I\} \cup \{ l^{\aatom{}} \mid l\in I\}\\
\cup \{ \katom\phi \mid \phi\in \Katom(T)\cap \K(M')\} \\  \cup\{ \aatom\phi\mid \phi\in \Aatom(T)\cap \A(M)\}\\
 \cup \{ \neg \katom\phi\mid \phi\in \Katom(T)\text{ and } \phi\notin \K(M')\}
\\ \cup \{\neg \aatom\phi\mid \phi\in \Aatom(T)\text{ and } \phi\notin \A(M)\}\\
\cup  \bigcup_{\psi\in \Katom(T)\atop \psi\in \K(M')} \{ l^{\katom\psi} \mid l\in I\} \cup \bigcup_{\psi\in \Aatom(T)\atop \psi\in \A(M)} \{ l^{\aatom\psi}\mid l\in I\} \\
\cup  \bigcup_{\psi\in \Katom(T) \atop \psi\notin \K(M')} \left\{ l^{\katom\psi} \mid l\in I' \text{, $I'$ is a model of $\K(M')\cup \{\neg \psi\}$}\right\}\\
\cup  \bigcup_{\psi\in \Aatom(T) \atop \psi\notin \A(M)} \left\{ l^{\aatom\psi} \mid l\in I' \text{, $I'$ is a model of $\A(M) \cup \{\neg \psi\}$}\right\}.
\end{multline*}

\noindent From the proof of Proposition~\ref{prop:1}, $I^{*\prime}$ is a model of~$\formOne_T$, and
\begin{align*}
  I^{*\prime} \cap \{ \aatom\phi \mid \phi\in \Atom_\A(T)\} &= I^* \cap \{\aatom\phi \mid \phi\in \Atom_\A(T)\} \\
  I^{*\prime} \cap \{ \katom\phi \mid \phi\in \Atom_\K(T)\} &\subsetneq I^*\cap \{ \katom\phi \mid \phi\in \Atom_\K(T)\}
\end{align*}
This conflicts to the second condition in the proposition, then the assumption is not valid. So there does not exist another model $M'$ of $T$ such that $\A(M') = \A(M)$ and $\K(M')\subsetneq \K(M)$, thus $M$ is a GK model of $T$.
\hfill
\end{longproof}

\begin{longproof}
\textbf{ of Theorem~\ref{them}:}

$\Rightarrow$:
Let $M$ be a GK model of $T$. From Proposition~\ref{prop:2}, there exists a model $I^*$ of $\formTwo_T$ such that $\K(M)= \A(M) = \Th( \{\phi\mid \phi\in \Atom_\K(T) \text{ and } I^* \models \katom\phi\})$.
We can create a set $S$ of literals as $S = I^* \cup \{u, v\}\ \cup $
\begin{quote}
  $\{ p^* \mid \text{for each new atom $p^*$ occurring in $\trc(T^*)$}\}\ \cup$\\
  $\{ c_\phi \mid \phi\in \Atom_\K(T)\}\cup \{ \hat{p}\mid p\in \Atom\}$.
\end{quote}
Clearly, $S$ satisfies each rule in $\trlp(T)$. Now we want to prove that $S$ is an answer set of the program.

Assume that $S$ is not an answer set of $\trlp(T)$, then there exists another set $S'\subsetneq S$ such that $S'$ satisfies each rule in the reduct $\trlp(T)^S$.
Note that, $I^*\subseteq S'$, $u$ implies $\{ p^* \mid \text{for each new atom $p^*$ occurring in $\trc(T^*)$}\}\cup \{ c_\phi \mid \phi\in \Atom_\K(T)\}$ and $v$ implies $\{ \hat{p}\mid p\in \Atom\}$.
Then there are only two possible cases: $u\notin S'$ or $v\notin S'$.

Case 1: $u\notin S'$, then there exists a set
\begin{multline*}
T = S'\cap \big( \{ p^* \mid p^* \text{ is a new atom occurring in $\trc(T^*)$}\}\\ \cup \{\aatom\phi\mid \phi\in \Atom_\A(T)\}\big)
\end{multline*}
such that $T$ satisfies 
$\trc(T^*)$.
For each $\phi\in \Katom(T)$,
\begin{itemize}
\item by the rule $u\gets c_\phi,\Not \katom\phi$, $I^* \models \neg \katom\phi$ implies $c_\phi\notin S'$;
\item by the rule $u\gets k^*_\phi, \Not \katom\phi$, $I^* \models \neg \katom\phi$ implies $k^*_\phi\notin S'$;
\item by rules $u\gets c_{\phi}, k^*_\phi, \Not \neg \katom\phi$ and $u; c_\phi; k^*_\phi \gets \Not \neg \katom\phi$, $I^* \models \katom\phi$ implies either $c_\phi$ or $k^*_\phi$ is in $S'$ but not both;
\item by the rule $u; c_{\phi_1}; \cdots; c_{\phi_m}\gets \top$, there exists $c_\psi \in S'$ for some $\psi\in \Katom(T)$.
\end{itemize}
So there exists $\psi\in \Katom(T)$ such that $\katom\psi\in S'$, $c_\psi\in S'$ and $k^*_\psi\notin S'$. Then
we could create an interpretation $I^{*\prime}$ as
\begin{multline*}
I^{*\prime} = \{ p\mid p\in \Atom \text{ and } p^*\in S'\}\\ \cup \{ \neg p \mid p\in\Atom \text{ and } p^*\notin S'\}\\
 \cup\{ \katom\phi \mid \phi\in \Atom_\K(T) \text{ and } k^*_\phi\in S'\}\\ \cup \{ \neg \katom\phi \mid \phi\in \Atom_\K(T)\text{ and } k^*_\phi\notin S'\} \\
\cup \{ \aatom\phi \mid \phi\in \Atom_\A(T)\text{ and } \aatom\phi\in S'\}\\ \cup \{\neg \aatom\phi \mid \phi\in \Atom_\A(T)\text{ and } \aatom\phi\notin S'\} \\
\cup \bigcup_{\psi\in \Katom(T)} \{ p^{\katom\psi} \mid p^{\katom\psi*} \in S'\} \cup \bigcup_{\psi\in \Katom(T)} \{ \neg p^{\katom\psi} \mid p^{\katom\psi*}\notin S'\}\\
\cup \bigcup_{\psi\in \Aatom(T)} \{ p^{\aatom\psi} \mid p^{\aatom\psi*} \in S'\} \cup \bigcup_{\psi\in \Aatom(T)} \{\neg p^{\aatom\psi} \mid p^{\aatom\psi*}\notin S'\}.
\end{multline*}
Clearly, $I^{*\prime}$ is a model of $\formTwo_T$. From the above results,
\begin{itemize}
  \item $I^{*\prime} \cap \{ \aatom\phi\mid \phi\in \Atom_\A(T)\} = I^{*}\cap \{\aatom\phi\mid \phi\in \Atom_\A(T)\}$, and
  \item $I^{*\prime} \cap \{\katom\phi\mid \phi\in \Atom_\K(T)\} \subsetneq I^*\cap \{\katom\phi\mid \phi\in \Atom_\K(T)\}$.
\end{itemize}
From Proposition~\ref{prop:2}, such $I^{*\prime}$ does not exist. This conflicts to the assumption, then Case 1 is impossible.

Case 2: $v\notin S'$, then there exists a set
\begin{multline*}
U = S'\cap \big( \{ \hat{a} \mid a\in \Atom\}\cup \{ \katom\phi\mid \phi\in \Atom_\K(T)\}\\\cup \{ \aatom\phi\mid \phi\in \Atom_\A(T)\}\big)
\end{multline*}
such that $U$ satisfies each rule in \\
$\trc(\bigwedge_{\phi\in \Atom_\K(T)} (\katom\phi\supset \widehat{\phi}) \land \neg \bigwedge_{\psi\in \Atom_\A(T)}(\aatom\psi\supset\widehat{\psi}))$.

Then there exists $\psi\in \Atom_\A(T)$ such that $I^*\models \aatom\psi$ and there exists an interpretation $I\subseteq \Lit$ such that $I\models \bigwedge_{\phi\in \Atom_\K(T), I^*\models \katom\phi} \phi \land \neg \psi$, thus $\psi \notin \Th(\{\phi\mid \phi\in\Atom_\K(T)\text{ and } I^*\models \katom\phi\})$. From Proposition~\ref{prop:2}, such $\psi$ does not exist. This conflicts to the assumption, then Case 2 is impossible.
So both cases are impossible, then $S'$ does not exist and $S$ is an answer set of $\trlp(T)$.

$\Leftarrow$: Let $S$ be an answer set of $\trlp(T)$. We can create an interpretation $I^*$ as
the intersection of $S$ with the set of atoms occurring in~$\formTwo_T$.
Clearly, $I^*$ is a model of $\formTwo_T$.

Similar to the above proof: 
If there exists another model $I^{*\prime}$ of $\formTwo_T$ such that
\begin{align*}
  I^{*\prime} \cap \{ \aatom\phi\mid \phi\in \Atom_\A(T)\} &= I^{*}\cap \{\aatom\phi\mid \phi\in \Atom_\A(T)\} \\
  I^{*\prime} \cap \{\katom\phi\mid \phi\in \Atom_\K(T)\} &\subsetneq I^*\cap \{\katom\phi\mid \phi\in \Atom_\K(T)\}
\end{align*}
then there exists another set $S'$ such that $S'$ satisfies each rule in the reduct $\trlp(T)^S$ and $u\notin S'$, thus $S'\subsetneq S$. This conflicts to the precondition that $S$ is an answer set, then such a model $I^{*\prime}$ does not exist.

If there exists $\psi\in \Atom_\A(T)$ such that $I^*\models \aatom\psi$ and $\psi\notin \Th(\{\phi\mid \phi\in \Atom_\K(T) \text{ and } I^*\models \katom\phi\})$, then there exists another set $S'$ such that $S'$ satisfies each rule in the reduct $\trlp(T)^S$ and $v\notin S'$, thus $S'\subsetneq S$. This conflicts to the precondition that $S$ is an answer set, then such $\psi$ does not exist.

From Proposition~\ref{prop:2}, a Kripke interpretation $M$ such that $\K(M)=\A(M) = \Th(\{\phi\mid \phi\in \Atom_\K(T) \text{ and } \katom\phi\in S\})$ is a GK models of $T$.
\hfill
\end{longproof}


\begin{thebibliography}{}

\bibitem[\protect\citeauthoryear{Baumann \bgroup et al.\egroup
  }{2010}]{baumann10statedefaults}
Baumann, R.; Brewka, G.; Strass, H.; Thielscher, M.; and Zaslawski, V.
\newblock 2010.
\newblock {State Defaults and Ramifications in the Unifying Action Calculus}.
\newblock In {\em KR},  435--444.

\bibitem[\protect\citeauthoryear{Bouveret and
  Lang}{2008}]{bouveret08efficiency}
Bouveret, S., and Lang, J.
\newblock 2008.
\newblock Efficiency and envy-freeness in fair division of indivisible goods:
  {L}ogical representation and complexity.
\newblock {\em JAIR} 32:525--564.

\bibitem[\protect\citeauthoryear{Brewka}{1994}]{brewka94addingpriorities}
Brewka, G.
\newblock 1994.
\newblock {Adding Priorities and Specificity to Default Logic}.
\newblock In {\em JELIA},  247--260.

\bibitem[\protect\citeauthoryear{Cadoli, Eiter, and
  Gottlob}{1994}]{cadoli94defaultlogic}
Cadoli, M.; Eiter, T.; and Gottlob, G.
\newblock 1994.
\newblock Default logic as a query language.
\newblock In {\em KR},  99--108.

\bibitem[\protect\citeauthoryear{Chen \bgroup et al.\egroup
  }{2010}]{chen10dl2asp}
Chen, Y.; Wan, H.; Zhang, Y.; and Zhou, Y.
\newblock 2010.
\newblock {dl2asp: Implementing Default Logic via Answer Set Programming}.
\newblock In {\em JELIA}, volume 6341,  104--116.

\bibitem[\protect\citeauthoryear{Cholewi{\'n}ski \bgroup et al.\egroup
  }{1999}]{cholewinski99computing}
Cholewi{\'n}ski, P.; Marek, V.~W.; Truszczy{\'n}ski, M.; and Mikitiuk, A.
\newblock 1999.
\newblock Computing with default logic.
\newblock {\em AIJ} 112(1):105--146.

\bibitem[\protect\citeauthoryear{Delgrande and
  Schaub}{2000}]{delgrande-schaub00expressingpreferences}
Delgrande, J.~P., and Schaub, T.
\newblock 2000.
\newblock {Expressing Preferences in Default Logic}.
\newblock {\em AIJ} 123(1--2):41--87.

\bibitem[\protect\citeauthoryear{Denecker, Marek, and
  Truszczy{\'n}ski}{2003}]{denecker03uniformsemantic}
Denecker, M.; Marek, V.~W.; and Truszczy{\'n}ski, M.
\newblock 2003.
\newblock {Uniform Semantic Treatment of Default and Autoepistemic Logics}.
\newblock {\em AIJ} 143(1):79--122.

\bibitem[\protect\citeauthoryear{Dix, Furbach, and
  Niemel{\"a}}{2001}]{dix01nonmonotonic}
Dix, J.; Furbach, U.; and Niemel{\"a}, I.
\newblock 2001.
\newblock Nonmonotonic reasoning: {T}owards efficient calculi and
  implementations.
\newblock {\em Handbook of Automated Reasoning} 2(18):1121--1234.

\bibitem[\protect\citeauthoryear{Drescher \bgroup et al.\egroup
  }{2008}]{claspD}
Drescher, C.; Gebser, M.; Grote, T.; Kaufmann, B.; K{\"o}nig, A.; Ostrowski,
  M.; and Schaub, T.
\newblock 2008.
\newblock {Conflict-Driven Disjunctive Answer Set Solving}.
\newblock In {\em KR},  422--432.

\bibitem[\protect\citeauthoryear{Eiter and
  Gottlob}{1995}]{eiter-gottlob95dasp-complexity}
Eiter, T., and Gottlob, G.
\newblock 1995.
\newblock On the computational cost of disjunctive logic programming:
  Propositional case.
\newblock {\em AMAI} 15(3--4):289--323.

\bibitem[\protect\citeauthoryear{Ferraris}{2005}]{Ferraris2005}
Ferraris, P.
\newblock 2005.
\newblock Answer sets for propositional theories.
\newblock In {\em LPNMR},  119--131.

\bibitem[\protect\citeauthoryear{Gebser, Kaufmann, and
  Schaub}{2013}]{gebser13claspD2}
Gebser, M.; Kaufmann, B.; and Schaub, T.
\newblock 2013.
\newblock Advanced conflict-driven disjunctive answer set solving.
\newblock In {\em IJCAI}.

\bibitem[\protect\citeauthoryear{Giunchiglia, Lierler, and
  Maratea}{2006}]{cmodels}
Giunchiglia, E.; Lierler, Y.; and Maratea, M.
\newblock 2006.
\newblock {Answer Set Programming Based on Propositional Satisfiability}.
\newblock {\em J. Autom. Reasoning} 36(4):345--377.

\bibitem[\protect\citeauthoryear{Heyting}{1930}]{heyting30ht}
Heyting, A.
\newblock 1930.
\newblock {Die formalen Regeln der intuitionistischen Logik}.
\newblock In {\em Sitzungsberichte der preußischen Akademie der
  Wissenschaften},  42--65, 57--71, 158--169.
\newblock Physikalisch-mathematische Klasse.

\bibitem[\protect\citeauthoryear{Janhunen and Niemel{\"a}}{2004}]{janhunen-gnt}
Janhunen, T., and Niemel{\"a}, I.
\newblock 2004.
\newblock {{G}n{T} -- A Solver for Disjunctive Logic Programs}.
\newblock In {\em LPNMR},  331--335.

\bibitem[\protect\citeauthoryear{Janhunen and
  Oikarinen}{2004}]{janhunen2004capturing}
Janhunen, T., and Oikarinen, E.
\newblock 2004.
\newblock Capturing parallel circumscription with disjunctive logic programs.
\newblock In {\em Logics in Artificial Intelligence}.
\newblock  134--146.

\bibitem[\protect\citeauthoryear{Ji and Lin}{2012}]{ji12}
Ji, J., and Lin, F.
\newblock 2012.
\newblock {From Turner's Logic of Universal Causation to the Logic of GK}.
\newblock In {\em Correct Reasoning}, volume 7265,  380--385.

\bibitem[\protect\citeauthoryear{Ji and Lin}{2013}]{ji-lin13turners}
Ji, J., and Lin, F.
\newblock 2013.
\newblock Turner's logic of universal causation, propositional logic, and logic
  programming.
\newblock In {\em LPNMR},  401--413.

\bibitem[\protect\citeauthoryear{Junker and
  Konolige}{1990}]{junker-konolige90computing}
Junker, U., and Konolige, K.
\newblock 1990.
\newblock {Computing the Extensions of Autoepistemic and Default Logics with a
  Truth Maintenance System}.
\newblock In {\em AAAI},  278--283.

\bibitem[\protect\citeauthoryear{Konolige}{1988}]{konolige88ontherelation}
Konolige, K.
\newblock 1988.
\newblock {On the Relation Between Default and Autoepistemic Logic}.
\newblock {\em AIJ} 35(3):343--382.

\bibitem[\protect\citeauthoryear{Lakemeyer and
  Levesque}{2005}]{lakemeyer05onlyknowing}
Lakemeyer, G., and Levesque, H.~J.
\newblock 2005.
\newblock Only-knowing: {T}aking it beyond autoepistemic reasoning.
\newblock In {\em AAAI},  633--638.

\bibitem[\protect\citeauthoryear{Leone \bgroup et al.\egroup }{2006}]{dlv}
Leone, N.; Pfeifer, G.; Faber, W.; Eiter, T.; Gottlob, G.; Perri, S.; and
  Scarcello, F.
\newblock 2006.
\newblock {The DLV system for knowledge representation and reasoning}.
\newblock {\em ACM Transactions on Computational Logic} 7(3):499--562.

\bibitem[\protect\citeauthoryear{Lifschitz, Tang, and Turner}{1999}]{nested99}
Lifschitz, V.; Tang, L.~R.; and Turner, H.
\newblock 1999.
\newblock {Nested expressions in logic programs}.
\newblock {\em AMAI} 25(3-4):369--389.

\bibitem[\protect\citeauthoryear{Lifschitz}{1994}]{lifschitz94mbnf}
Lifschitz, V.
\newblock 1994.
\newblock Minimal belief and negation as failure.
\newblock {\em AIJ} 70(1--2):53--72.

\bibitem[\protect\citeauthoryear{Lin and Shoham}{1992}]{gk}
Lin, F., and Shoham, Y.
\newblock 1992.
\newblock {A logic of knowledge and justified assumptions}.
\newblock {\em AIJ} 57(2-3):271--289.

\bibitem[\protect\citeauthoryear{Lin and Zhou}{2011}]{LinZhou11}
Lin, F., and Zhou, Y.
\newblock 2011.
\newblock From answer set logic programming to circumscription via logic of
  {GK}.
\newblock {\em AIJ} 175(1):264--277.

\bibitem[\protect\citeauthoryear{Lin}{2002}]{Lin:kr02}
Lin, F.
\newblock 2002.
\newblock Reducing strong equivalence of logic programs to entailment in
  classical propositional logic.
\newblock In {\em KR},  170--176.

\bibitem[\protect\citeauthoryear{Marek and
  Truszczy{\'n}ski}{1991}]{marek91computing}
Marek, V.~W., and Truszczy{\'n}ski, M.
\newblock 1991.
\newblock Computing intersection of autoepistemic expansions.
\newblock In {\em LPNMR},  37--50.

\bibitem[\protect\citeauthoryear{Martin and
  Thielscher}{2001}]{martin01addressing}
Martin, Y., and Thielscher, M.
\newblock 2001.
\newblock {Addressing the Qualification Problem in FLUX}.
\newblock In {\em KI/{\"O}GAI},  290--304.

\bibitem[\protect\citeauthoryear{McCarthy}{1980}]{McCarthy80}
McCarthy, J.
\newblock 1980.
\newblock Circumscription -- a form of non-monotonic reasoning.
\newblock {\em AIJ} 13:295--323.

\bibitem[\protect\citeauthoryear{McCarthy}{1986}]{McCarthy86}
McCarthy, J.
\newblock 1986.
\newblock Applications of circumscription to formalizing commonsense knowledge.
\newblock {\em AIJ} 28:89--118.

\bibitem[\protect\citeauthoryear{Moore}{1985}]{Moore}
Moore, R.
\newblock 1985.
\newblock Semantical considerations on nonmonotonic logic.
\newblock {\em AIJ} 25(1):75--94.

\bibitem[\protect\citeauthoryear{Nicolas, Saubion, and
  St{\'e}phan}{2000}]{nicolas00gadel}
Nicolas, P.; Saubion, F.; and St{\'e}phan, I.
\newblock 2000.
\newblock Gadel: a genetic algorithm to compute default logic extensions.
\newblock In {\em ECAI},  484--490.

\bibitem[\protect\citeauthoryear{Niemel{\"a}}{1995}]{niemela95decision}
Niemel{\"a}, I.
\newblock 1995.
\newblock A decision method for nonmonotonic reasoning based on autoepistemic
  reasoning.
\newblock {\em J.\ Autom.\ Reasoning} 14(1):3--42.

\bibitem[\protect\citeauthoryear{Pagnucco \bgroup et al.\egroup
  }{2013}]{pagnucco13implementing}
Pagnucco, M.; Rajaratnam, D.; Strass, H.; and Thielscher, M.
\newblock 2013.
\newblock {Implementing Belief Change in the Situation Calculus and an
  Application}.
\newblock In {\em LPNMR}, volume 8148,  439--451.

\bibitem[\protect\citeauthoryear{Reiter}{1980}]{reiter80}
Reiter, R.
\newblock 1980.
\newblock A logic for default reasoning.
\newblock {\em AIJ} 13(1-2):81--132.

\bibitem[\protect\citeauthoryear{Reiter}{1987}]{reiter87diagnosis}
Reiter, R.
\newblock 1987.
\newblock A theory of diagnosis from first principles.
\newblock {\em AIJ} 32(1):57--95.

\bibitem[\protect\citeauthoryear{Risch and Schwind}{1994}]{risch94tableaux}
Risch, V., and Schwind, C.
\newblock 1994.
\newblock Tableaux-based characterization and theorem proving for default
  logic.
\newblock {\em J.\ Autom.\ Reasoning} 13(2):223--242.

\bibitem[\protect\citeauthoryear{Rosati}{1999}]{rosati99reasoning}
Rosati, R.
\newblock 1999.
\newblock Reasoning about minimal belief and negation as failure.
\newblock {\em JAIR} 11:277--300.

\bibitem[\protect\citeauthoryear{Schaub and Nicolas}{1997}]{schaub97xray}
Schaub, T., and Nicolas, P.
\newblock 1997.
\newblock An implementation platform for query-answering in default logics:
  {T}he {XRay} system, its implementation and evaluation.
\newblock In {\em LPNMR}.
\newblock  441--452.

\bibitem[\protect\citeauthoryear{Schwarz and
  Truszczynski}{1994}]{schwarz-truszczynski94minimal-knowledge}
Schwarz, G., and Truszczynski, M.
\newblock 1994.
\newblock Minimal knowledge problem: {A} new approach.
\newblock {\em AIJ} 67(1):113--141.

\bibitem[\protect\citeauthoryear{Strass}{2012}]{strass12draculasp}
Strass, H.
\newblock 2012.
\newblock The draculasp system: {D}efault reasoning about actions and change
  using logic and answer set programming.
\newblock In {\em NMR}.

\bibitem[\protect\citeauthoryear{Thielscher}{1996}]{thielscher96causalityand}
Thielscher, M.
\newblock 1996.
\newblock {Causality and the Qualification Problem}.
\newblock In {\em KR},  51--62.

\bibitem[\protect\citeauthoryear{Truszczy{\'n}ski}{2007}]{truszczynski07s4f}
Truszczy{\'n}ski, M.
\newblock 2007.
\newblock The modal logic {S4F}, the default logic, and the logic
  here-and-there.
\newblock In {\em AAAI},  508--514.

\bibitem[\protect\citeauthoryear{Turner}{1999}]{Turner}
Turner, H.
\newblock 1999.
\newblock {Logic of universal causation}.
\newblock {\em AIJ} 113(1):87--123.

\end{thebibliography}
\end{document}